# Plane Constraints Aided Multi-Vehicle Cooperative Positioning Using Factor Graph Optimization

Chen Zhuang, Hongbo Zhao

*Abstract*—The development of vehicle-to-vehicle (V2V) communication facilitates the study of cooperative positioning (CP) techniques for vehicular applications. The CP methods can improve the positioning availability and accuracy by inter-vehicle ranging and data exchange between vehicles. However, the inter-vehicle ranging can be easily interrupted due to many factors such as obstacles in-between two cars. Without inter-vehicle ranging, the other cooperative data such as vehicle positions will be wasted, leading to performance degradation of range-based CP methods. To fully utilize the cooperative data and mitigate the impact of inter-vehicle ranging loss, a novel cooperative positioning method aided by plane constraints is proposed in this paper. The positioning results received from cooperative vehicles are used to construct the road plane for each vehicle. The plane parameters are then introduced into CP scheme to impose constraints on positioning solutions. The state-of-art factor graph optimization (FGO) algorithm is employed to integrate the plane constraints with raw data of Global Navigation Satellite Systems (GNSS) as well as inter-vehicle ranging measurements. The proposed CP method has the ability to resist the interruptions of inter-vehicle ranging since the plane constraints are computed by just using position-related data. A vehicle can still benefit from the position data of cooperative vehicles even if the inter-vehicle ranging is unavailable. The experimental results indicate the superiority of the proposed CP method in positioning performance over the existing methods, especially when the inter-ranging interruptions occur.

*Index Terms*—Cooperative positioning (CP), inter-vehicle ranging, plane constraints, factor graph optimization (FGO).

## I. INTRODUCTION

PROVIDING accurate and reliable positioning service is essential for vehicular applications in intelligent transportation systems (ITS) [1], such as lane-level guidance, flexible lane management, formation driving and self-driving. The Global Navigation Satellite Systems (GNSS), which have become the core enabler of the vehicular positioning systems, are capable of providing users with absolute positions in open-sky areas. Unfortunately, the performance of standalone GNSS can be degraded severely due to GNSS non-line-of-sight (NLOS) delays, multipath effects and signal outages in dense urban areas [2]. To improve the positioning performance in urban scenes, many onboard sensors such as inertial navigation system (INS) [3], Lidar [4], and camera [5] are introduced into vehicular navigation system. These sensor-rich vehicles (SRVs) can get benefit from high-performance sensors by applying advanced sensor fusion methods. However, it is of great difficulty for common vehicles (CoVs) equipped with few low-cost perception sensors to obtain high-precision positions by fusing their own measurements [6]. In fact, CoVs will still coexist with SRVs in the near future. It is worth studying how to improve the positioning performance of CoVs, especially in dense urban areas.

Owing to the rapid development of vehicle-to-vehicle (V2V) communication, cooperative positioning (CP) methods also receive the widespread attention recent years [7]. In a CP system, the vehicles can exchange their measuring data with each other for improving their positioning performance [8]. Since a vehicle is feasible to conduct the cooperative positioning without utilizing expensive sensors except for inter-vehicle communication devices, CP becomes one of the most promising positioning methods in the case of coexistence of CoVs and SRVs. The inter-node ranging is a requisite for most of the existing CP methods. Based on different techniques for inter-node ranging, the CP methods can be divided into two categories, including GNSS-based CP methods and range-based CP methods [9]. The GNSS-based CP methods mainly apply the double difference (DD) on the pseudorange [10, 11] or carrier phase measurements [12, 13] to obtain inter-node distances or relative positions between two vehicles. Although GNSS DD techniques have been used by many researchers for inter-node ranging in CP methods, the availability and accuracy of double differenced GNSS is limited in dense urban environments due to multipath effects, NLOS delays and signal outages [14, 15]. Different from GNSS-based CP methods, the range-based CP methods estimate the inter-node ranges by using radio ranging techniques such as time-of arrival (TOA) [16], angle of arrival (AOA) [17], receiver signal strength (RSS) [18] and round-trip time (RTT) [19]. The range-based CP methods are more suitable for the scenarios where the availability of GNSS is limited [20]. Additional radio ranging measurements could help to improve the positioning availability and decrease the positioning errors.

Chen Zhuang, Hongbo Zhao, Shan Hu and Wenquan Feng are with School of Electronic and Information Engineering, Beihang University, Beijing 100101, China (e-mail: zhuangchen0214@buaa.edu.cn, bhzhb@buaa.edu.cn, hushan@buaa.edu.cn, buaafwq@buaa.edu.cn).

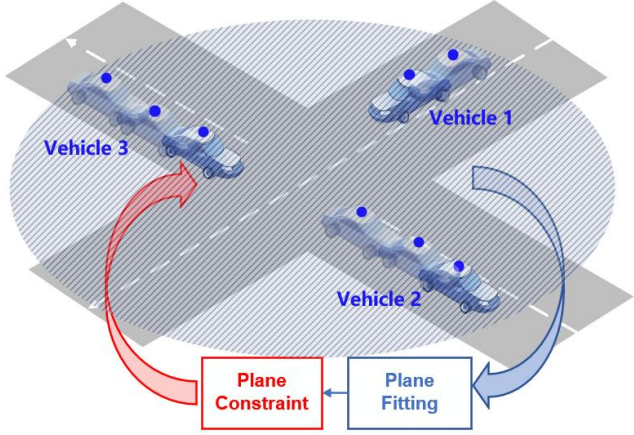

Fig. 1. Illustration of the road plane construction by just using the position-related data of vehicles. The blue spots denote the positioning results of different vehicles in multiple epochs, which will be used for fitting road plane.

A variety of range-based CP methods have been proposed in the latest literatures [21-25]. Most of these works assume that the inter-node ranging is always available and accurate enough. However, the inter-node ranging can be easily affected by many factors such as NLOS delays, obstacles in-between two cars and limited perception range. Besides, unstable vehicular communication also degrades the performance of range-based CP methods. The excessively large transmission latency, packet collisions and limited link life time could reduce the availability and reliability of inter-node ranging. Some researchers have mentioned the problem of abnormal inter-node ranging and ranging data loss in range-based CP methods. To solve the problem of abnormal inter-node ranging, some fault-tolerant CP methods are proposed based on fault detection and exclusion (FDE) algorithm [26-28]. However, the availability of inter-node ranging will decline severely if numerous abnormal measurements are removed. To mitigate the influence of ranging data loss on CP methods, some works focus on optimizing the communication protocols and recovering the missing inter-node distances. The authors in [29] analyzed the impact of range information exchange overhead on CP methods and proposed protocol improvements which can reduce the packet collisions and improve the positioning accuracy. Nevertheless, protocol improvements can only solve the problem of ranging data loss brought by communication congestion. In [30], the missing inter-node distances are estimated by applying the singular value thresholding so that the impact of inadequate inter-node ranging can be mitigated. However, the accumulated errors are inevitable for inter-node distance estimation if a large proportion of inter-node ranges are missing. To sum up, it is still questionable whether the CP methods can benefit from the shared information in the case that most of the inter-ranging measurements are interrupted or even unavailable.

In addition to the ranging information, a positioning method can also benefit from the range-free environmental data such as geographic information. The road conditions have already been considered in some non-cooperative positioning methods [31, 32]. Considering that the urban roads are relatively smooth and the road conditions are satisfying, a vehicle can be regarded to travel on a plane most of the time. If the road plane information is available, the plane constraints can be imposed on the positioning solutions to improve the positioning performance. Fortunately, the road plane can be constructed by just using the position-related data from the cooperative vehicles that have ever traveled on this plane, as shown in Fig. 1. Therefore, it is possible to use the position solutions shared among the vehicles for CP purposes.

In this paper, we aim at using the plane constraints to enhance the performance of the range-based CP method. The state-of-art factor graph optimization (FGO) algorithm is employed to fuse all the measurements in CP system. The FGO algorithm is suitable for applications with multiple constraints. There have been some works that apply FGO to GNSS positioning and cooperative positioning [33-35]. The related works show the superiority of FGO over the extended Kalman filter in positioning estimation. The main contributions of this paper are listed as follows:

1) We propose to use the positioning results received from the cooperative vehicles to construct the road plane for each vehicle, and then introduce the plane parameters into the CP scheme. In this way, a vehicle can still benefit from the position data of cooperative vehicles even if the inter-node ranging is unavailable.

2) We design the plane availability detection and fault exclusion method to ensure the effectiveness of the plane constraints. The outliers can be removed from the plane fitting by using Random Sample Consensus (RANSAC) algorithm.

3) Based on the FGO algorithm, we propose a novel centralized framework to integrate the plane constraints with the GNSS pseudorange, Doppler measurements and inter-vehicle ranging measurements (if available).

The paper is organized as follows. In Section 2, we give an overview of the proposed method. In section 3, the plane construction method is described, including the plane detection and outlier exclusion algorithm. In Section 4, the CP method based on FGO is introduced in detail. In Section 5, we present and analyze the experimental results from different aspects. The conclusions are given in the end of the paper.

## II. OVERVIEW OF THE PROPOSED METHOD

The overview of the proposed CP method aided by plane constraints is shown in Fig. 2. It is assumed that the vehicular network is composed by a set of nodes with cardinality $M$. For each node, a road plane where the vehicle node is traveling is constructed locally based on the cooperative positioning results calculated in the past. The positioning solutions from all the nodes in the vehicular network can be used to construct the road planes. The plane parameters may be different among the nodes because the vehicle nodes are possible to be located on different road planes. A vehicle node will collect the plane parameters together with the raw measurements including GNSS pseudorange, GNSS Doppler and inter-node distance to prepare the node message. All the node messages will be sent to center server for cooperative positioning. The global states of all the nodes are estimated centralized based on the state-of-art FGO algorithm. The positioning results of all the vehicles are collected and fed back to each vehicle for the road plane construction.

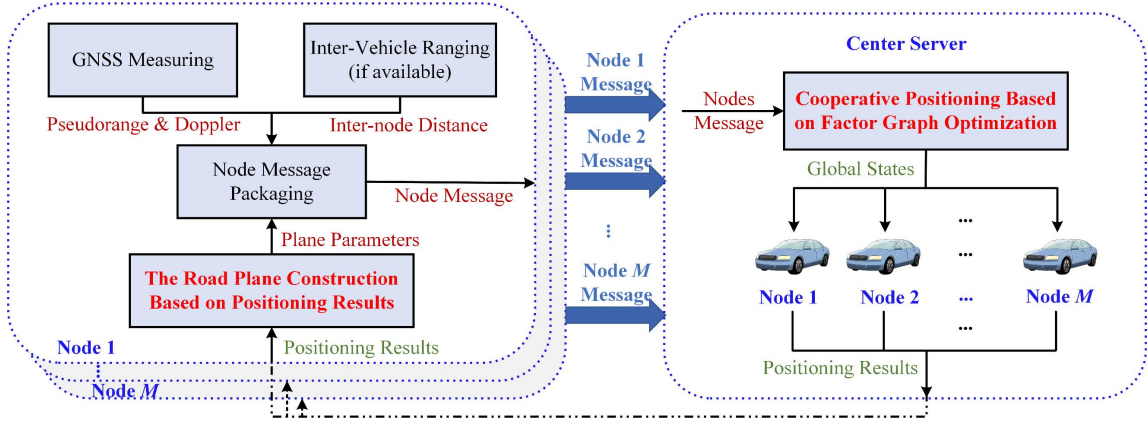

Fig. 2. The framework of the proposed cooperative positioning method aided by plane constraints.

The innovation of our work is to introduce the plane constraints into CP method and then study how to integrate the plane constraints with other measurements using FGO, so as to improve the performance of the CP methods in the case of unstable inter-node ranging. On the one hand, the proposed CP method is able to benefit from the position information of cooperative vehicles even if the inter-vehicle ranging is unavailable. On the other hand, the use of plane constraints can further improve the positioning accuracy when the inter-node ranging is available. The details about the plane construction and the FGO-based CP algorithm will be introduced in the following sections.

III. THE ROAD PLANE CONSTRUCTION BASED ON POSITIONING RESULTS

The road plane construction is the key of the proposed method. The procedure of the road plane construction is shown in Fig. 3. Firstly, the plane fitting based on SVD is conducted by fusing a series of positioning results from both local and cooperative vehicles. Then, the availability of the fitted plane is determined via the plane detection. If the plane detection is passed, the plane parameters will be output for constraining the positions. Otherwise, a fault searching strategy based on RANSAC will be used to search and exclude the outliers in the plane fitting. After that, the plane fitting and the plane detection will be conducted once again. The plane will be marked as unavailable if the plane detection fails to be passed after the fault exclusion. The plane fitting, the plane detection and the fault exclusion method will be introduced as follows.

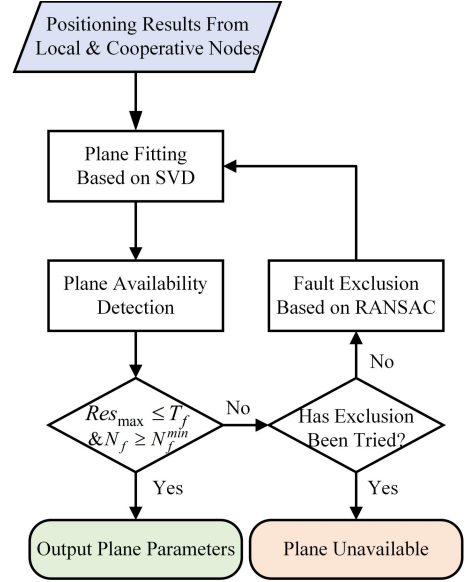

Fig. 3. The structure of road plane construction based on positioning results.

*3.1 Plane fitting*

The positioning results of the vehicles are related to the location of GNSS antennas, which are usually placed on the top of the vehicles. Since the heights of vehicles are different, the heights of GNSS antennas from the ground are also different from each other. Before fitting the road planes, we need to eliminate the height differences of GNSS antennas among the vehicles. The projection of the positioning results on the road surface will be used to fit the plane instead of the positioning results themselves.

For a positioning result $\tilde{\mathbf{p}}_{u_m,t}$ of the node $u_m$ at epoch $t$ in Earth-Centered-Earth-Fixed (ECEF) frame, the projection of this position on the road plane can be calculated by:

$$\tilde{\mathbf{p}}'_{u_m,t} = \tilde{\mathbf{p}}_{u_m,t} - \mathbf{S}\begin{bmatrix}0;0;h_{u_m}\end{bmatrix} \quad (1)$$

where $\mathbf{S}$ represents the transformation matrix from East-North-Up (ENU) to ECEF coordinate system. $h_{u_m}$ is the antenna height of $u_m$. We use $\tilde{\mathbf{p}}'_{u_m,t} = \begin{bmatrix}\tilde{x}'_{u_m,t}, \tilde{y}'_{u_m,t}, \tilde{z}'_{u_m,t}\end{bmatrix}^T$ to fit the planes in the follows.

The positioning results of all the vehicles in the past epochs are collected together. As introduced in the previous section, a total of $M$ nodes is assumed to be considered. The set of the positioning results from the beginning to epoch ($t$-1) for all the vehicles is formulated as follows:

$$pos^t = \{\tilde{\mathbf{p}}'_{u_1,0}, \tilde{\mathbf{p}}'_{u_1,1} \ldots, \tilde{\mathbf{p}}'_{u_1,t-1}, \ldots, \tilde{\mathbf{p}}'_{u_M,0}, \tilde{\mathbf{p}}'_{u_M,1} \ldots, \tilde{\mathbf{p}}'_{u_M,t-1}\} \quad (2)$$

By default, we will fit a plane for each vehicle at each measuring epoch. For a specific vehicle, we select the positioning results close to this vehicle to fit the plane. The point set used to fit the plane for $u_m$ at epoch $t$ is defined as:

$$\begin{aligned} Set^{u_m,t}_{fitting} &= \left\{\forall \tilde{\mathbf{p}}' \in pos^t \mid \|\tilde{\mathbf{p}}' - \tilde{\mathbf{p}}^{SPP}_{u_m,t}\| < range_{fitting}\right\} \\ &= \left\{\tilde{\mathbf{p}}'^{u_m,t}_{fitting,1}, \tilde{\mathbf{p}}'^{u_m,t}_{fitting,2}, \ldots, \tilde{\mathbf{p}}'^{u_m,t}_{fitting,n}, \ldots \tilde{\mathbf{p}}'^{u_m,t}_{fitting,N_f}\right\} \end{aligned} \quad (3)$$

where $\tilde{\mathbf{p}}^{SPP}_{u_m,t}$ is the raw position calculated by GNSS single point positioning (SPP). $range_{fitting}$ denotes the range for selecting the fitting points, which is set to 50 meters in our research. $\tilde{\mathbf{p}}'^{u_m,t}_{fitting,n} = \left[\tilde{x}'^{u_m,t}_{fitting,n}, \tilde{y}'^{u_m,t}_{fitting,n}, \tilde{z}'^{u_m,t}_{fitting,n}\right]^T$ represents a fitting point in $Set^{u_m,t}_{fitting}$. $N_f$ denotes the initial number of fitting points in this set. It is noted that the maximum value of $N_f$ is set to 20 in this paper so as to reduce the computational burden of plane fitting. If there are more than 20 fitting points in $Set^{u_m,t}_{fitting}$, the fitting points will be removed randomly from the set until the number of fitting points is equal to 20.

In this paper, the least squares (LS) method is employed to fit the road plane $\alpha$ for $u_m$ at epoch $t$. The equation of this plane can be expressed as:

$$\alpha: P^A_{u_m,t} x + P^B_{u_m,t} y + P^C_{u_m,t} z + P^D_{u_m,t} = 0 \quad (4)$$

where $P^A_{u_m,t}$, $P^B_{u_m,t}$, $P^C_{u_m,t}$ and $P^D_{u_m,t}$ are the plane parameters to be estimated. Here, we impose a constraint on the normal vector of the fitted plane as follows:

$$\left(P^A_{u_m,t}\right)^2 + \left(P^B_{u_m,t}\right)^2 + \left(P^C_{u_m,t}\right)^2 = 1 \quad (5)$$

We calculate the center of the fitting points in $Set^{u_m,t}_{fitting}$ by:

$$\begin{aligned} \overline{\mathbf{p}}^{u_m,t}_{fitting} &= \left[\frac{1}{N_f}\sum_{n=1}^{N_f} \tilde{x}'^{u_m,t}_{fitting,n}, \frac{1}{N_f}\sum_{n=1}^{N_f} \tilde{y}'^{u_m,t}_{fitting,n}, \frac{1}{N_f}\sum_{n=1}^{N_f} \tilde{z}'^{u_m,t}_{fitting,n}\right] \\ &= \left[\overline{x}^{u_m,t}_{fitting}, \overline{y}^{u_m,t}_{fitting}, \overline{z}^{u_m,t}_{fitting}\right]^T \end{aligned}$$
(6)

where $\overline{\mathbf{p}}^{u_m,t}_{fitting}$ should also be on the plane. Then, we can derive the following equations:

$$P^A_{u_m,t} \times \left(\tilde{x}'^{u_m,t}_{fitting,n} - \overline{x}^{u_m,t}_{fitting}\right) + P^B_{u_m,t} \times \left(\tilde{y}'^{u_m,t}_{fitting,n} - \overline{y}^{u_m,t}_{fitting}\right) + P^C_{u_m,t} \times \left(\tilde{z}'^{u_m,t}_{fitting,n} - \overline{z}^{u_m,t}_{fitting}\right) = 0$$
(7)

Considering that there are $N_f$ fitting points, we extend the equation (7) to the follows:

$$\mathbf{A}\mathbf{X}^{u_m,t}_P = \mathbf{0}$$

$$\text{with } \mathbf{A} = \begin{bmatrix} \left(\tilde{\mathbf{p}}'^{u_m,t}_{fitting,1}\right)^T - \left(\overline{\mathbf{p}}^{u_m,t}_{fitting}\right)^T \\ \left(\tilde{\mathbf{p}}'^{u_m,t}_{fitting,2}\right)^T - \left(\overline{\mathbf{p}}^{u_m,t}_{fitting}\right)^T \\ \vdots \\ \left(\tilde{\mathbf{p}}'^{u_m,t}_{fitting,N_f}\right)^T - \left(\overline{\mathbf{p}}^{u_m,t}_{fitting}\right)^T \end{bmatrix}_{N_f \times 3} \quad (8)$$

where $\mathbf{X}^{u_m,t}_P = \left[P^A_{u_m,t}, P^B_{u_m,t}, P^C_{u_m,t}\right]^T$ is the normal vector to be estimated. We utilize the singular value decomposition (SVD) method to calculate the normal vector. Here, $\mathbf{A}$ can be decomposed into:

$$\mathbf{A} = \mathbf{U\Sigma V}^T \quad (9)$$

where $\mathbf{U}$ and $\mathbf{V}$ belong to unitary matrices. $\mathbf{\Sigma}$ is a diagonal matrix in which the diagonal elements are singular values. It is assumed that the last diagonal element is the minimum singular value. The eigenvector associated with the minimum singular value corresponds to the normal vector, which can be formulated as follows:

$$\begin{cases} \tilde{P}^A_{u_m,t} = \mathbf{V}(1,3) \\ \tilde{P}^B_{u_m,t} = \mathbf{V}(2,3) \\ \tilde{P}^C_{u_m,t} = \mathbf{V}(3,3) \end{cases} \quad (10)$$

where $\left[\tilde{P}^A_{u_m,t}, \tilde{P}^B_{u_m,t}, \tilde{P}^C_{u_m,t}\right]^T$ is the estimated normal vector. The remaining term of the plane equation can be calculated by:

$$\tilde{P}^D_{u_m,t} = -\left[\tilde{P}^A_{u_m,t}, \tilde{P}^B_{u_m,t}, \tilde{P}^C_{u_m,t}\right]^T \cdot \overline{\mathbf{p}}^{u_m,t}_{fitting} \quad (11)$$

In order to use the planes to constrain the positioning results, it is necessary to translate the road planes to the location of GNSS antennas. For vehicle $u_m$, a plane parallel to the fitted plane $\alpha$ can be expressed as:

$$\beta: \tilde{P}^A_{u_m,t} x + \tilde{P}^B_{u_m,t} y + \tilde{P}^C_{u_m,t} z + \tilde{P}'^D_{u_m,t} = 0 \quad (12)$$

where,

$$\tilde{P}'^D_{u_m,t} = \begin{cases} \tilde{P}^D_{u_m,t} + h_{u_m}, & \tilde{P}^D_{u_m,t} \geq 0 \\ \tilde{P}^D_{u_m,t} - h_{u_m}, & \tilde{P}^D_{u_m,t} < 0 \end{cases} \quad (13)$$

The distance between these two planes is equal to the antenna height of the target vehicle. It is noted that we will employ the plane $\beta$ instead of the plane $\alpha$ to constrain the positioning results.

*3.2 Plane Availability Detection and Fault Exclusion*

Not all the fitted planes can be used for constraining the positioning solutions. There are some requirements that should be met before using the plane constraints. Firstly, a flat road without curved surface is an essential prerequisite for applying plane constraints. If a vehicle travels on a road with curved surface, it is impossible to fit a plane where the vehicle is truly located. Secondly, the positioning results used to fit the planes should be accurate enough. Some positioning solutions with excessively large errors (i.e. outliers) can lead to an inaccurate fitted plane. The fitted plane may be far away from the true road plane in this case. Therefore, it is necessary to determine whether the plane constraints are effective or not by applying the plane availability detection. A fault exclusion procedure is also needed to remove the potential outliers from plane fitting.

**Algorithm 1** Fault Exclusion Based on RANSAC

**Inputs:** The set $Set_{fitting}^{u_m,t}$ with $N_f$ fitting points, the maximum iteration number $I_{max}$, the threshold for selecting inliers $T_f$.

1. Initialize a point set $Set_{RANSAC}^{u_m,t} \leftarrow \emptyset$ and the number of the elements in this set $N'_f \leftarrow 0$;
2. **for** $iteration \leftarrow 1$ to $I_{max}$ **do**
3.   Initialize the set of inliers $Set_{inlier} \leftarrow \emptyset$ and the number of inliers $N_{in} \leftarrow 0$;
4.   Compute a plane $\alpha_1$ based on three fitting points selected randomly from $Set_{fitting}^{u_m,t}$;
5.   **for** $n \leftarrow 1$ to $N_f$ **do**
6.     $D'_n \leftarrow$ Calculate the distance between a fitting point $\tilde{\mathbf{p}}_{fitting,n}^{\prime u_m,t}$ and the plane $\alpha_1$ according to (15);
7.     **if** $D'_n < T_f$ **then**
8.       $N_{in} \leftarrow N_{in} + 1$; // Update the number of inliers
9.       Add the current inlier $\tilde{\mathbf{p}}_{fitting,n}^{\prime u_m,t}$ to $Set_{inlier}$;
10.    **end if**
11.  **end for**
12.  **if** $N_{in} > N'_f$ **then**
13.    $N'_f \leftarrow N_{in}$;
14.    $Set_{RANSAC}^{u_m,t} \leftarrow Set_{inlier}$; // Save the set that has more inliers
15.  **end if**
16.  **if** $N'_f = N_f$ **then**
17.    **break**; // Stop when all the fitting points are inliers
18.  **end if**
19. **end for**
20. $N_f \leftarrow N'_f$; // Update the number of the final fitting points

**Outputs:** The set $Set_{RANSAC}^{u_m,t}$ with the most inliers.

In this paper, the availability of the fitted planes is mainly determined by the residuals in the plane fitting. The maximum value of the residuals can be calculated as follows:

$$Res_{max} = \max\{D_1, D_2, \ldots, D_n, \ldots, D_{N_f}\} \quad (14)$$

with

$$D_n = \frac{\left|\tilde{P}_{u_m,t}^A \times \tilde{x}_{fitting,n}^{\prime u_m,t} + \tilde{P}_{u_m,t}^B \times \tilde{y}_{fitting,n}^{\prime u_m,t} + \tilde{P}_{u_m,t}^C \times \tilde{z}_{fitting,n}^{\prime u_m,t} + \tilde{P}_{u_m,t}^D\right|}{\sqrt{\left(\tilde{P}_{u_m,t}^A\right)^2 + \left(\tilde{P}_{u_m,t}^B\right)^2 + \left(\tilde{P}_{u_m,t}^C\right)^2}}$$

(15)

where $D_n$ represents the distance between a fitting point and the fitted plane $\alpha$, i.e. residual. The maximum residual $Res_{max}$ will be relatively small if all the fitting points are accurate enough and located on a flat road without curved surface. Otherwise, the residuals might be relatively large. The plane availability can be determined by comparing the residuals of plane fitting with a threshold. To further ensure the reliability of the plane fitting and the following fault exclusion procedure, the final number of the fitting points has to be large enough. Therefore, whether a fitted plane is available or not can be determined by the following equation:

$$\begin{cases} Res_{max} \leq T_f, N_f \geq N_f^{min} & Plane\ Available \\ Res_{max} \leq T_f, N_f < N_f^{min} & Plane\ Unavailable \\ Res_{max} > T_f, N_f \geq N_f^{min} & Plane\ Unavailable \\ Res_{max} > T_f, N_f < N_f^{min} & Plane\ Unavailable \end{cases} \quad (16)$$

where $T_f$ denotes the threshold related to the fitting residuals. The value of $T_f$ is set to 5 meters in our research. $N_f^{min}$ represents the minimum number of the fitting points to ensure the reliability of the plane fitting and the following fault exclusion, which is set to 15 in this paper. Considering that the maximum number of fitting points is set to 20 initially, up to 5 outliers can be excluded by performing the following fault exclusion.

To improve the availability of the plane constraints, a fault exclusion procedure is conducted to remove the potential outliers from the fitting data set if the plane availability detection fails to be passed. The Random Sample Consensus (RANSAC) algorithm is employed to exclude the outliers in the plane fitting, which is well known in the field of graphics and image processing. This algorithm calculates an estimate based on the minimal necessary subset of the fitting points so that the number of outliers employed in the estimation can be minimized [36]. Firstly, we compute a plane based on three fitting points selected randomly from the fitting data set. Then, the distances between the remaining fitting points and this plane are calculated and compared with a threshold. Here, we use the same threshold $T_f$ in equation (16) for selecting the inliers. If the distance between a fitting point and the plane is less than $T_f$, this point is regarded as an inlier. Otherwise, the point will be marked as an outlier. The above steps will be repeated until the maximum number of iterations is reached or there is no outlier. We will count the number of inliers for each iteration to find the one with the most inliers. More inliers indicate a higher consensus of the current plane with the remaining fitting points. Finally, we output the corresponding point set with the most inliers, which will be used to fit the plane again. It is noted that the number of inliers must be large enough to ensure the reliability of RANSAC algorithm. The details of the RANSAC-based fault exclusion method can be seen in **Algorithm 1**. The maximum iteration number is set to 7 by experience, so as to keep a balance between the fault exclusion ability and computational load.

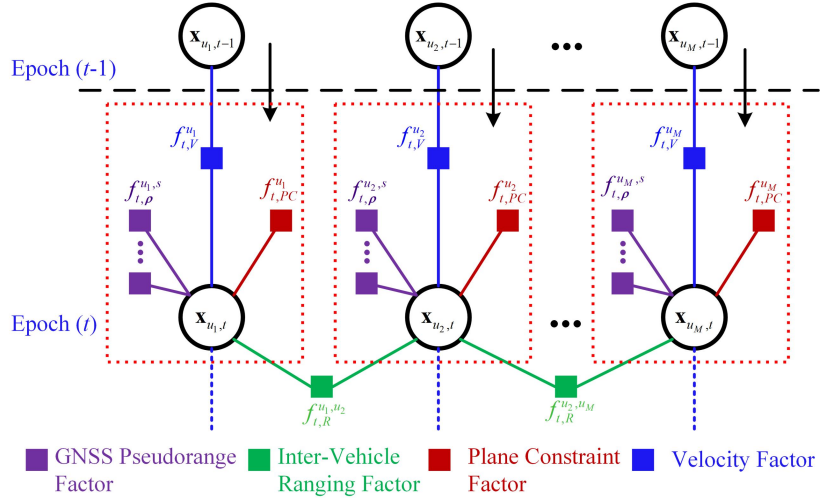

Fig. 4. The structure of cooperative positioning algorithm based on FGO.

## IV. COOPERATIVE POSITIONING ALGORITHM BASED ON FACTOR GRAPH OPTIMIZATION

The structure of the proposed factor graph for solving the cooperative positioning is shown in Fig. 4. The dashed boxes represent physical nodes, i.e. vehicles. Time is simply denoted by the discrete epoch index $t$. Referring to a particular node $u_m$, the state of the node at epoch $t$ is defined as:

$$\mathbf{x}_{u_m,t} = \left[ \mathbf{p}_{u_m,t}^T, b_{u_m,t} \right]^T \quad (17)$$

where $\mathbf{p}_{u_m,t} = \left[ x_{u_m,t}, y_{u_m,t}, z_{u_m,t} \right]^T$ is the position of the node in ECEF frame. $b_{u_m,t}$ represents the clock bias of the node in distance units.

Within the factor graph of a single node, the purple and red rectangles represent GNSS and the plane constraint, respectively. It is noted that the plane data obtained from Section III will be used to calculate the plane constraint factors in this section. The blue rectangles denote the velocity factors which connect two consecutive epochs inside a node. The green rectangles connecting two nodes denote the inter-vehicle ranging factors. If the inter-ranging measurements are available, the inter-vehicle ranging factors will also be considered in the FGO-based positioning algorithm.

The optimal positioning solution of each node can be obtained by finding a configuration of node states that matches all the factors as much as possible. To reduce the computation load, only the factors inside a sliding window will be utilized in FGO. The rest of this section presents the details about how to calculate the factors mentioned above and how to optimize the states by integrating these factors.

### 4.1 GNSS Pseudorange Factor

For a particular node $u_m$, the GNSS pseudorange measurement between the node and the satellite $s$ can be expressed as:

$$\rho_{u_m,t}^s = r_{u_m,t}^s + b_{u_m,t} - b_t^s + I_t^s + T_t^s + \varepsilon_{\rho,u_m}^s \quad (18)$$

where $r_{u_m,t}^s = \left\| \mathbf{p}_{u_m,t} - \mathbf{p}_{s,t} \right\|$ represents the geometric distance between the node $u_m$ and the satellite $s$. $\mathbf{p}_{s,t} = \left[ x_{s,t}, y_{s,t}, z_{s,t} \right]^T$ is the position of the satellite $s$ at epoch $t$ in ECEF frame. The symbol $\|\cdot\|$ represents the $l_2$ norm operation. $b_t^s$ is the satellite clock bias in distance units. $I_t^s$ and $T_t^s$ denote the ionospheric and tropospheric delays, respectively. $\varepsilon_{\rho,u_m}^s$ is the error caused by receiver noise and multipath effects. The differential GNSS (DGNSS) corrections are applied to mitigate the ionospheric, tropospheric, satellite clock bias and orbit errors, respectively. Thus, the observation model for pseudorange measurements can be written as:

$$\rho_{u_m,t}^s - \delta\rho_{DGNSS,t}^s = h_{t,\rho}^{u_m,s}\left( \mathbf{p}_{u_m,t}, \mathbf{p}_{s,t}, b_{u_m,t} \right) + \varepsilon_{\rho,u_m}^s \quad (19)$$

$$h_{t,\rho}^{u_m,s}\left( \mathbf{p}_{u_m,t}, \mathbf{p}_{s,t}, b_{u_m,t} \right) = \left\| \mathbf{p}_{u_m,t} - \mathbf{p}_{s,t} \right\| + b_{u_m,t} \quad (20)$$

where $\delta\rho_{DGNSS,t}^s$ is the DGNSS corrections of the satellite $s$ at epoch $t$. The error function for a single GNSS pseudorange factor can be calculated as follows:

$$f_{t,\rho}^{u_m,s} = \rho_{u_m,t}^s - \delta\rho_{DGNSS,t}^s - h_{t,\rho}^{u_m,s}\left( \mathbf{p}_{u_m,t}, \mathbf{p}_{s,t}, b_{u_m,t} \right) \quad (21)$$

It is noted that an integrity monitoring algorithm will be used to detect and exclude the faulty measurements with large errors [28].

### 4.2 Plane Constraint Factor

In section III, the positioning results of cooperative vehicles are used to fit the road planes where the vehicles are traveling. If the fitted planes are available and accurate enough, we can use these planes to constrain the positioning solutions. For a given node $u_m$, the plane constraint model can be expressed as:

$$P_{u_m,t}^{\prime D} = h_{t,PC}^{u_m}\left( \mathbf{p}_{u_m,t}, \mathbf{u}_{u_m,t}^{PC} \right) + \varepsilon_{PC} \quad (22)$$

$$h_{t,PC}^{u_m}\left(\mathbf{p}_{u_m,t},\mathbf{u}_{u_m,t}^{PC}\right) = \mathbf{p}_{u_m,t} \cdot \mathbf{u}_{u_m,t}^{PC}$$
$$= -\tilde{P}_{u_m,t}^{A} \times x_{u_m,t} - \tilde{P}_{u_m,t}^{B} \times y_{u_m,t} - \tilde{P}_{u_m,t}^{C} \times z_{u_m,t} \quad (23)$$

where $\mathbf{u}_{u_m,t}^{PC} = \left[-\tilde{P}_{u_m,t}^{A}, -\tilde{P}_{u_m,t}^{B}, -\tilde{P}_{u_m,t}^{C}\right]^T$ is the unit normal vector of the fitted plane. $\tilde{P}_{u_m,t}^{A}$, $\tilde{P}_{u_m,t}^{B}$, $\tilde{P}_{u_m,t}^{C}$, and $\tilde{P}_{u_m,t}^{\prime D}$ can be estimated in section III. $\varepsilon_{PC}$ is the residual in the plane fitting. Therefore, we calculate the error function for a plane constraint factor as follows:

$$f_{t,PC}^{u_m} = P_{u_m,t}^{\prime D} - h_{t,PC}^{u_m}\left(\mathbf{p}_{u_m,t},\mathbf{u}_{u_m,t}^{PC}\right) \quad (24)$$

By default, the plane constraint factors will be calculated for each vehicle once a new epoch is added to the factor graph. To reduce the computational burden on vehicles, we can also use the same plane to constrain the positioning solutions as long as the variation of plane parameters within the sliding window is small enough.

### 4.3 Inter-vehicle Ranging Factor

The inter-vehicle ranging can be performed by the round-trip-time (RTT) method, which avoids the clock bias between two nodes. For the node $u_m$ and the node $u_n$, the inter-vehicle ranging model can be expressed as:

$$d_t^{u_m,u_n} = h_{t,R}^{u_m,u_n}\left(\mathbf{p}_{u_m,t},\mathbf{p}_{u_n,t}\right) + \varepsilon_d \quad (25)$$

$$h_{t,R}^{u_m,u_n}\left(\mathbf{p}_{u_m,t},\mathbf{p}_{u_n,t}\right) = \left\|\mathbf{p}_{u_m,t} - \mathbf{p}_{u_n,t}\right\| \quad (26)$$

where $d_t^{u_m,u_n}$ denotes the inter-node ranging measurement. $\varepsilon_d$ is the additive white Gaussian noise variables for inter-node ranging. Thus, the error function for an inter-vehicle ranging factor can be obtained as follows:

$$f_{t,R}^{u_m,u_n} = d_t^{u_m,u_n} - h_{t,R}^{u_m,u_n}\left(\mathbf{p}_{u_m,t},\mathbf{p}_{u_n,t}\right) \quad (27)$$

It is noted that the inter-vehicle ranging factor would be excluded from the objective function of the FGO if the inter-node ranging is unavailable.

### 4.4 Velocity Factor

The velocity factors connect two consecutive states of a node. Since GNSS Doppler measurements are less sensitive to the multipath effects, we use Doppler measurements to estimate the velocity of a vehicle node. The relationship between a GNSS Doppler measurement and the velocity can be expressed as follows:

$$-\lambda f_{d,u_m,t}^s = \left(\mathbf{v}_{s,t} - \mathbf{v}_{u_m,t}\right) \cdot \mathbf{e}_{u_m,t}^s + \delta f_{u_m,t} - \delta f_t^s + \varepsilon_{f_d,u_m}^s$$
$$\Rightarrow -\lambda f_{d,u_m,t}^s - \mathbf{v}_{s,t} \cdot \mathbf{e}_{u_m,t}^s + \delta f_t^s = -\mathbf{v}_{u_m,t} \cdot \mathbf{e}_{u_m,t}^s + \delta f_{u_m,t} \quad (28)$$

where $f_{d,u_m,t}^s$ is GNSS Doppler measurement for satellite $s$ and $\lambda$ represents the wavelength of the signal. $\mathbf{v}_{u_m,t} = \left[v_{x,t}^{u_m}, v_{y,t}^{u_m}, v_{z,t}^{u_m}\right]^T$ is the velocity of the node and $\delta f_{u_m,t}$ stands for the clock drift of the node, both of which are unknowns in velocity estimation. $\mathbf{v}_{s,t}$ denotes the velocity of the satellite and $\delta f_t^s$ is the clock drift of the satellite, which can be calculated based on the broadcast ephemeris. $\varepsilon_{f_d,u_m}^s$ represents the residual errors. $\mathbf{e}_{u_m,t}^s$ is the unit direction vector from the node to the satellite, which is formulated as:

$$\mathbf{e}_{u_m,t}^s = \left[e_{x,u_m,t}^s, e_{y,u_m,t}^s, e_{y,u_m,t}^s\right]^T$$
$$= \left[\frac{x_{s,t} - \hat{x}_{u_m,t}}{\left\|\tilde{\mathbf{p}}_{u_m,t}^{SPP} - \mathbf{p}_{s,t}\right\|}, \frac{y_{s,t} - \hat{y}_{u_m,t}}{\left\|\tilde{\mathbf{p}}_{u_m,t}^{SPP} - \mathbf{p}_{s,t}\right\|}, \frac{z_{s,t} - \hat{z}_{u_m,t}}{\left\|\tilde{\mathbf{p}}_{u_m,t}^{SPP} - \mathbf{p}_{s,t}\right\|}\right]^T \quad (29)$$

where $\tilde{\mathbf{p}}_{u_m,t}^{SPP} = \left[\tilde{x}_{u_m,t}^{SPP}, \tilde{y}_{u_m,t}^{SPP}, \tilde{z}_{u_m,t}^{SPP}\right]^T$ is the raw position calculated by GNSS SPP.

According to equation (28), we can obtain the observation model for Doppler measurements as follows:

$$\mathbf{y}_{d,u_m,t} = \mathbf{G}\begin{bmatrix} v_{x,t}^{u_m} \\ v_{y,t}^{u_m} \\ v_{z,t}^{u_m} \\ \delta f_{u_m,t} \end{bmatrix} + \boldsymbol{\varepsilon}_{f_d,u_m} \quad (30)$$

where,

$$\mathbf{y}_{d,u_m,t} = \left[y_{d,u_m,t}^{s_1}, y_{d,u_m,t}^{s_2}, y_{d,u_m,t}^{s_3}, \ldots\right]^T \quad (31)$$

$$y_{d,u_m,t}^s = -\lambda f_{d,u_m,t}^s - \mathbf{v}_{s,t} \cdot \mathbf{e}_{u_m,t}^s + \delta f_t^s \quad (32)$$

$$\mathbf{G} = \begin{bmatrix} -e_{x,u_m,t}^{s_1} & -e_{y,u_m,t}^{s_1} & -e_{z,u_m,t}^{s_1} & 1 \\ -e_{x,u_m,t}^{s_2} & -e_{y,u_m,t}^{s_2} & -e_{z,u_m,t}^{s_2} & 1 \\ -e_{x,u_m,t}^{s_3} & -e_{y,u_m,t}^{s_3} & -e_{z,u_m,t}^{s_3} & 1 \\ \vdots & \vdots & \vdots & \vdots \end{bmatrix} \quad (33)$$

Therefore, the velocity of the vehicle node can be estimated by using the least squares (LS) method as follows:

$$\begin{bmatrix} v_{x,t}^{u_m} \\ v_{y,t}^{u_m} \\ v_{z,t}^{u_m} \\ \delta f_{u_m,t} \end{bmatrix} = \left(\mathbf{G}^T\mathbf{G}\right)^{-1}\mathbf{G}^T\mathbf{y}_{d,u_m,t} \quad (34)$$

Then, the observation model for the velocity is formulated as:

$$\mathbf{v}_{u_m,t} = h_{t,V}^{u_m}\left(\mathbf{p}_{u_m,t},\mathbf{p}_{u_m,t-1}\right) + \boldsymbol{\varepsilon}_{V,u_m}$$
$$= \begin{bmatrix} \left(x_{u_m,t} - x_{u_m,t-1}\right)/\Delta t \\ \left(y_{u_m,t} - y_{u_m,t-1}\right)/\Delta t \\ \left(z_{u_m,t} - z_{u_m,t-1}\right)/\Delta t \end{bmatrix} + \boldsymbol{\varepsilon}_{V,u_m} \quad (35)$$

where $\boldsymbol{\varepsilon}_{V,u_m}$ represents the noise vector of the velocity measurements. $\Delta t$ is the time difference between two consecutive epochs. The error function for a velocity factor is written as:

$$f_{t,V}^{u_m} = \mathbf{v}_{u_m,t} - h_{t,V}^{u_m}\left(\mathbf{p}_{u_m,t}, \mathbf{p}_{u_m,t-1}\right) \quad (36)$$

where the velocity is given by the estimation (34).

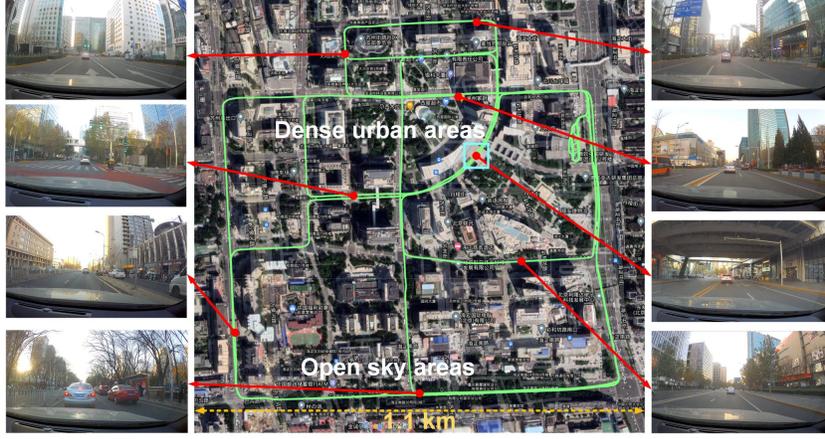

Fig. 5. The test routes and the driving situations.

*4.5 Pseudorange/Plane constraint/Inter-Ranging/Velocity Integration by FGO*

After the factors are derived, we can use them to estimate the states by solving the objective function. We formulate the objective function of the proposed FGO as follows:

$$\boldsymbol{\chi}^* = \arg\min_{\boldsymbol{\chi}} \sum_{m,s,t}\left\|f_{t,\rho}^{u_m,s}\right\|_{\boldsymbol{\Sigma}_{t,\rho}^{u_m,s}}^2 + \sum_{m,t}\left\|f_{t,PC}^{u_m}\right\|_{\boldsymbol{\Sigma}_{t,PC}^{u_m}}^2 \\ + \sum_{m,n,t}\left\|f_{t,R}^{u_m,u_n}\right\|_{\boldsymbol{\Sigma}_{t,R}^{u_m,u_n}}^2 + \sum_{m,t}\left\|f_{t,V}^{u_m}\right\|_{\boldsymbol{\Sigma}_{t,V}^{u_m}}^2 \quad (37)$$

where $\boldsymbol{\chi} = \left[\mathbf{X}_{u_1}^T, \mathbf{X}_{u_2}^T, \ldots, \mathbf{X}_{u_M}^T\right]^T$ is the global state of *M* nodes. $\mathbf{X}_{u_m} = \left[\mathbf{x}_{u_m,t_0}^T, \mathbf{x}_{u_m,t_0+1}^T, \ldots, \mathbf{x}_{u_m,t_0+l-1}^T\right]^T$ is the set of states of the node $u_m$ from the epoch $t_0$ to the current epoch ($t_0+l-1$). $l$ denotes the total epochs considered in the FGO, i.e. window length. $\boldsymbol{\chi}^*$ is the optimal estimation of the global state. $\boldsymbol{\Sigma}_{t,\rho}^{u_m,s}$ represents the noise covariance matrix of pseudorange factors, which is calculated based on carrier-to-noise ratio (CNR) of the signals [37]. The noise covariance matrix of the plane constraint factors is given as $\boldsymbol{\Sigma}_{t,PC}^{u_m}$, which is determined by the size of residual in plane fitting. $\boldsymbol{\Sigma}_{t,R}^{u_m,u_n}$ denotes the noise covariance matrix of the inter-node ranging factors, which depends on the inter-ranging device. Considering that UWB sensors are used in our research, the standard deviation of the inter-node ranging measurement is set to 0.3 m according to product data sheet [38]. $\boldsymbol{\Sigma}_{t,V}^{u_m}$ is the noise covariance matrix of the velocity factors associated with Doppler velocity measurements. $\boldsymbol{\Sigma}_{t,V}^{u_m}$ is set as $\sigma_V \mathbf{I}_{3\times 3}$. $\sigma_V$ is the scaling factor which is set to 0.6 according to [35].

Finally, we employ the Levenberg-Marquardt optimization method to solve the least squares optimization problem. Since the update time will climb linearly with increasing epochs, we delete the old states and their associated factors so as to reduce the computational load. The states and the factors are pruned depending on their age. All the states and factors that are older than 5 measuring cycles will be removed from the FGO.

V. EXPERIMENTS AND PERFORMANCE ANALYSIS

The proposed method is verified via the real dataset collected in the urban areas. Several experiments were carried out in Zhongguancun science park, Beijing, China. The test routes and the surrounding environments are depicted in Fig. 5. In order to apply GNSS differential corrections to the positioning results, a reference station was set in an open-sky area adjacent to the test routes. The Network Real Time Kinematic (NRTK) technique is employed to calculate the absolute position of the reference station.

*5.1 Experiment setup*

As depicted in Fig. 6, four vehicles were involved in the experiments, referred to as $u_1$, $u_2$, $u_3$ and $u_4$ in this section. All the vehicles traveled along the same routes during the experiments. There are two lanes in one direction and the vehicles traveled in different lanes most of the time. The vehicle $u_1$ was equipped with a GNSS receiver named M300 and a high-performance navigation system named NovAtel SPAN-ISA-100C. The other three vehicles were equipped with the same devices, including a GNSS receiver named OEM628 and a high-performance navigation system named NPOS220, as shown in Fig. 7. The GNSS measurements collected by M300 and OEM628 are used for algorithm evaluation, including pseudorange and Doppler measurements of GPS L1/L2 and Beidou B1/B2. The sampling rate is 1Hz for GNSS measurements. The satellite visibility of vehicle $u_1$ is shown in Fig. 8. It can be seen that the number of visible satellites decreases obviously in dense urban areas. Benefiting from geodetic-grade GNSS receivers with multiple antennas and high-precision inertial measuring units (IMU), the high-performance navigation systems can provide the reference trajectory for each vehicle. By the post-processing of the NovAtel Inertial Explorer software, the root mean squared error (RMSE) of the ground truth can reach 0.01m for horizontal direction and 0.02m for vertical direction.

All the vehicles were also equipped with Ultra-Wide Band (UWB) sensors to obtain the inter-node ranging measurements. Both GNSS receiver and UWB sensor were connected to a laptop which was used for data storage. The GNSS receivers can provide UWB sensors with precise GNSS time to synchronize GNSS and UWB measurements. The sampling rate is also 1Hz for UWB measurements. To reduce the system biases induced by the position difference between the GNSS antenna and the UWB antenna, all the UWB sensors were placed nearby the GNSS antennas. The maximum ranging distance for these UWB sensors is 500m. Since the vehicles kept close to each other during the experiments, the UWB measurements were available almost at all time. To simulate the losses of inter-node ranging measurements, we remove a proportion of the UWB measurements artificially in the following algorithm analysis.

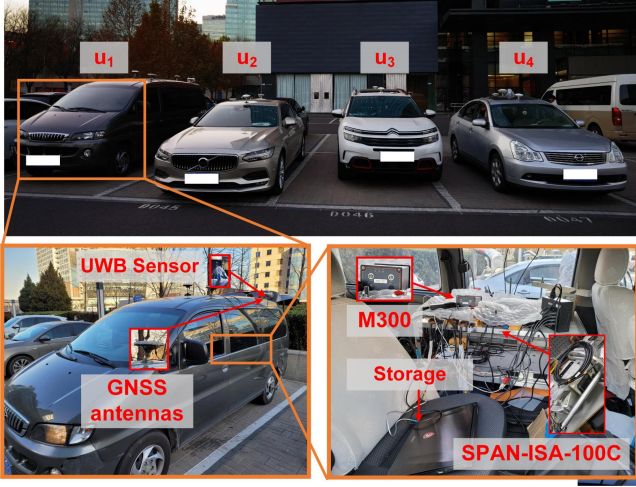

Fig. 6. The test vehicles and the devices equipped on vehicle $u_1$.

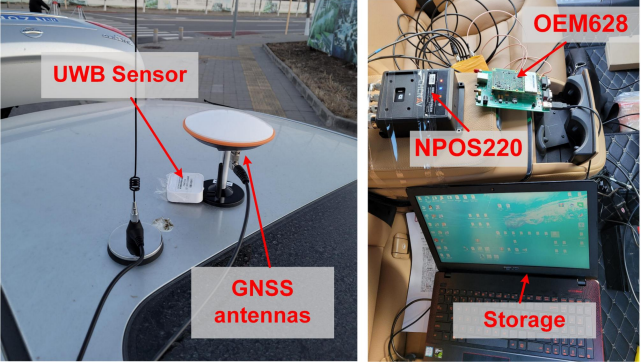

Fig. 7. The devices equipped on vehicle $u_4$. The same devices are equipped on the vehicle $u_2$ and $u_3$.

To obtain more reliable positioning results in urban areas, it is necessary to perform a fault detection and exclusion procedure before using the measurements. The cooperative integrity monitoring (CIM) method proposed in [28] is conducted to remove the GNSS faults and the UWB faults simultaneously. Most of the faulty measurements with large errors can be excluded by using CIM.

## 5.2 Comparison of Positioning Performance Using Different Methods

This section aims to demonstrate the effectiveness of the proposed CP method and compare the performance of the proposed method with the existing methods. Five control groups are considered in this section, which are presented as follows:

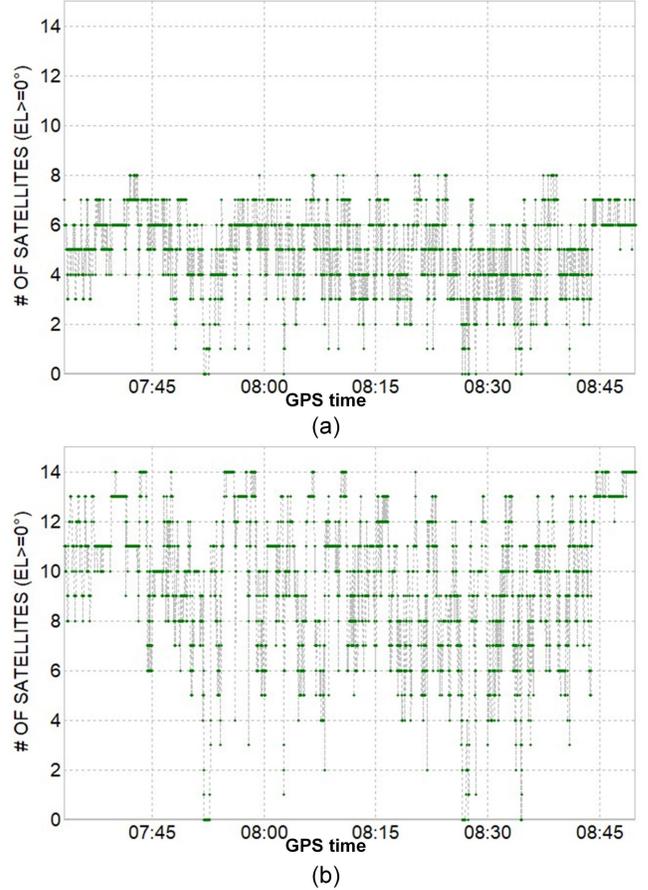

Fig. 8. The satellite visibility of vehicle $u_1$ during the tests for (a) GPS and (b) Beidou system.

- Non-cooperative positioning (Non-CP): The GNSS standalone positioning method based on extended Kalman filtering (EKF) is chosen as a non-CP method, which is compared with CP methods to demonstrate the benefit of cooperation. The pseudorange and Doppler measurements are used to estimate the states. A widely-used software named RTKLIB is employed to calculate GNSS standalone positioning results. The details about RTKLIB can be found in [39].
- GNSS-based CP method: The GNSS-based CP method applies the double difference (DD) on the shared pseudoranges to obtain relative positions between two vehicles, which are then combined with absolute positions to calculate final CP solutions. Here, the EKF is used for CP tight integration. The details about the GNSS-based CP method can be found in [10].
- Generalized approximate message passing (GAMP): This method can make full use of GNSS pseudoranges, Doppler shifts, inter-node ranging measurements and

4measurements of other sensors inside the CP system by applying GAMP. This method is based on belief propagation that employs the marginal distribution to estimate the states, which shows superior positioning performance than traditional Kalman filtering. Here, we only consider GNSS measurements and inter-node ranging measurements so as to make a fair comparison with other range-based CP methods. The details about GAMP-based CP method can be found in [19].

- Multi-Agent collaborative integration (MCI): This CP method can integrate the measurements from GNSS, inter-ranging sensors and other sensors using FGO algorithm. This method is chosen because it also applies FGO algorithm to the cooperative positioning. We only consider GNSS measurements and inter-ranging measurements so that we can make a fair comparison between MCI and other range-based CP methods. The details about MCI can be found in [34].
- Plane constraint (PC) aided CP: This is the proposed CP method, in which the plane constraints are introduced into the range-based CP scheme.

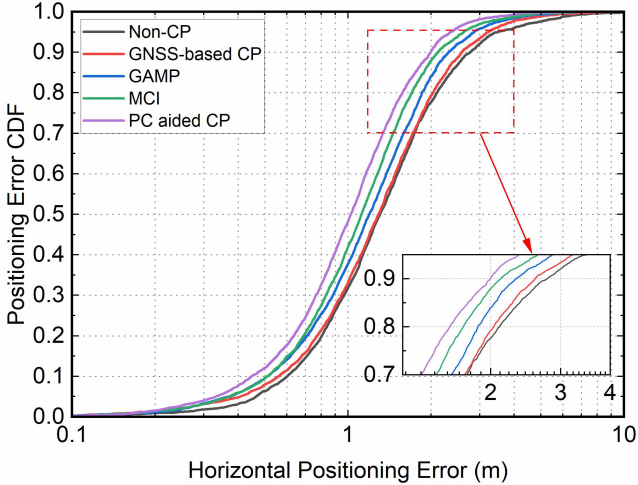

Fig. 9. The CDFs of the horizontal positioning errors for different methods.

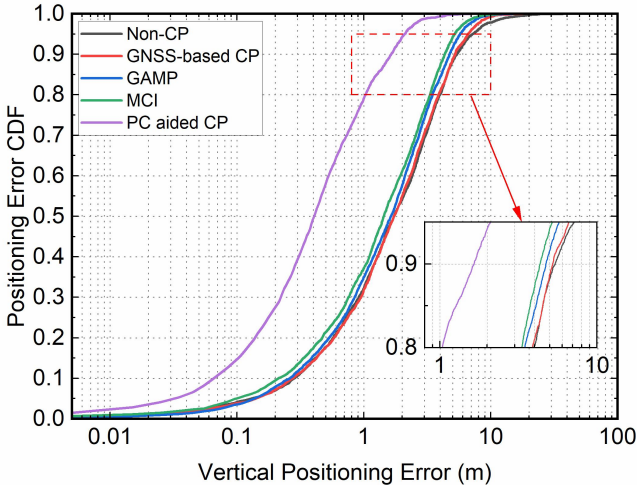

Fig. 10. The CDFs of the vertical positioning errors for different methods.

In this section, all the four vehicles are involved to calculate positioning results. Only the results of vehicle $u_1$ are presented to show the performance of different positioning methods. All the available UWB ranging measurements are used for range-based CP methods. We focus on both the horizontal and vertical positioning performance in the following analysis.

The cumulative distribution functions (CDF) of horizontal positioning errors (HPE) and vertical positioning errors (VPE) for vehicle $u_1$ under different positioning methods are depicted in Fig. 9 and Fig. 10. The statistic results are given in Table I, including horizontal root mean squared error (RMSE), vertical RMSE, circular error at 95% probable (CEP95), the maximum HPE and the maximum VPE. Without using any cooperative information, non-CP method shows the worst performance among all these positioning methods. The maximum HPE of non-CP method reaches 15.02 m due to the limited visibility of GNSS satellites and multipath effects in dense urban areas. All the CP methods can benefit from the shared data and show improvements on positioning accuracy. Compared to non-CP method, the GNSS-based CP method sees a slight decline in horizontal RMSE, with the figure decreasing from 1.98 m to 1.78 m. The improvement is not obvious because the inter-node ranges computed by GNSS DD pseudorange are not accurate enough. Although the correlated errors are eliminated through GNSS differencing, uncorrelated errors (multipath and receiver noise) increase instead, especially in urban areas. The GAMP method and MCI method show better performance than GNSS-based CP methods since these two methods utilize inter-node ranging measurements additionally. The positioning error of MCI method is smaller than GAMP method, with the figure of horizontal RMSE declining from 1.61 m to 1.5 m. The reason for this is that the FGO used by MCI method can take advantages of the historical measurements to further enhance the positioning performance.

The proposed CP method aided by plane constraints shows the highest positioning accuracy among these methods. Compared with MCI method, the horizontal RMSE of the proposed method is further reduced, reaching to 1.35 m for vehicle $u_1$. There is also a significant improvement on vertical positioning accuracy, with the figure of vertical RMSE dropping to 0.95 m. The plane constraints which can be regarded as new observations for each vehicle make a great contribution to performance improvements. It is noted that the vertical positioning errors are remarkably larger than the horizontal ones for non-CP, GNSS-based CP, GAMP and MCI method. However, for the proposed method, the positioning accuracy in vertical direction is significantly higher than that in horizontal direction. The performance gain is greater in the vertical direction although the performance improvements can be seen in both horizontal and vertical directions for the proposed method. Besides, the proposed method also shows a drop in the maximum HPE and VPE, with the figure decreasing to 7.65 m and 9.41 m, which demonstrates the reliability of the proposed method in dense urban areas.

*5.3 Performance Analysis Under Inter-Node Ranging Interruption*

To further demonstrate the superiority of the proposed method in the case that the inter-node ranging is limited, the inter-node ranging interruption is simulated by removing a proportion of inter-ranging measurements artificially. Specifically, we randomly remove 20%, 40%, 60%, 80% and 100% of inter-node ranging measurements between each pair of the nodes, respectively. The MCI method is chosen to make a comparison with the proposed method in this subsection.

TABLE I
STATISTICS FOR POSITIONING PERFORMANCE USING DIFFERENT METHODS

| Methods | Horizontal RMSE(m) | Vertical RMSE(m) | CEP95 (m) | Maximum HPE (m) | Maximum VPE (m) |
|---|---|---|---|---|---|
| Non-CP | 1.98 | 3.74 | 3.47 | 15.02 | 36.71 |
| GNSS-based CP | 1.78 | 3.21 | 3.22 | 10.11 | 19.70 |
| GAMP | 1.61 | 2.84 | 2.85 | 10.29 | 17.52 |
| MCI | 1.50 | 2.62 | 2.63 | 11.18 | 19.51 |
| PC aided CP | 1.35 | 0.95 | 2.36 | 7.65 | 9.41 |

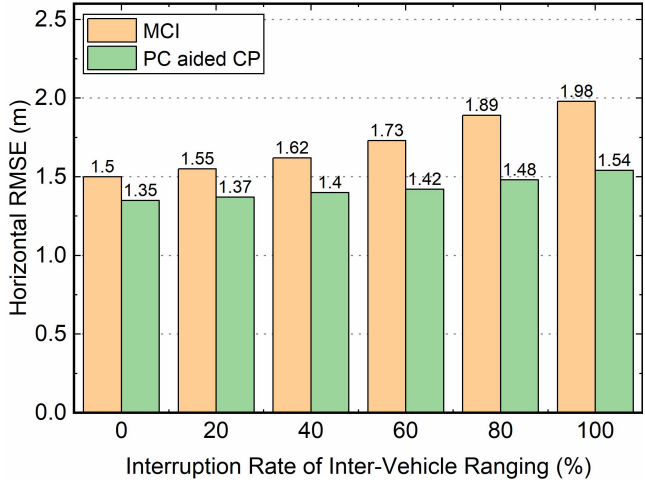

Fig. 11. The horizontal RMSE for the proposed method and MCI method under different interruption rates of inter-vehicle ranging.

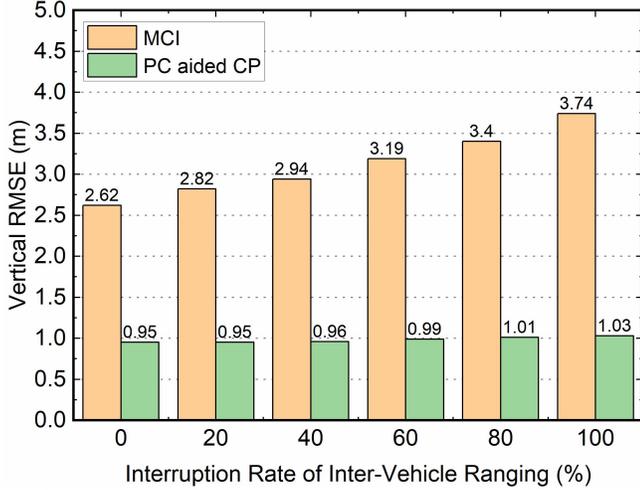

Fig. 12. The vertical RMSE for the proposed method and MCI method under different interruption rates of inter-vehicle ranging.

As shown in Fig. 11 and Fig. 12, the horizontal and vertical RMSE of vehicle $u_1$ using the proposed method and MCI method is compared under different inter-ranging interruption rates. The results without interruption (i.e. 0% case) are also given. The MCI method sees an upward trend in positioning errors with the increase of inter-ranging interruption rate. The horizontal RMSE of the MCI method grows remarkably, rising from 1.5 m for no interruption to 1.98 m for 100% interruption. Similar trend can be found in the vertical positioning results. The losses of inter-node ranging measurements lead to the descent of spatial correlation between the nodes, which degrades the positioning accuracy of range-based CP method severely. The positioning errors of the proposed method also increase with the growth of ranging interruption rate, with the figure of horizontal and vertical RMSE climbing to 1.48 m and 1.01 m in the case of 80% interruption rate, respectively. However, the performance loss caused by radio ranging interruption is obviously smaller for the proposed method. It can be seen that the differences between the positioning errors of the proposed method and MCI method increase with the rise of ranging interruption rate. The introduction of plane constraints can compensate the performance degradations induced by the losses of inter-ranging measurements to some extent. Compared with the results in Table I, it can be seen that the positioning accuracy of the proposed method is still higher than GNSS-based CP method and GAMP method although all the inter-ranging measurements are removed.

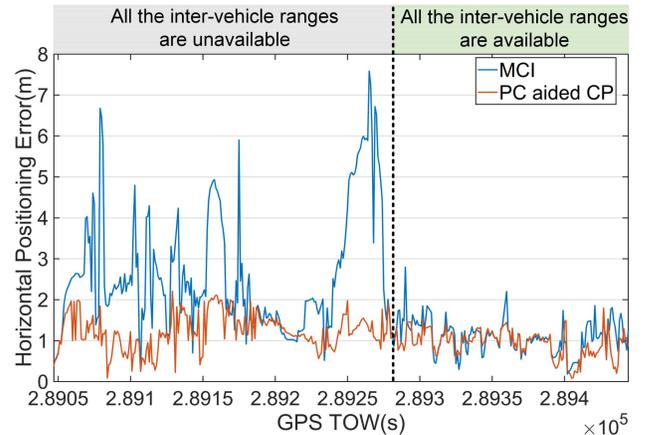

Fig. 13. The detailed horizontal positioning errors for the proposed method and MCI method from epoch 289045 to 289445. The periods when the inter-ranging interruptions occur are also marked.

To show the details of the positioning performance under inter-ranging interruption, we present the HPE and the inter-ranging availability of vehicle $u_1$ in Fig. 13. The horizontal axis represents the time of week (TOW) of GPS in seconds. Here, we select the positioning results from the epoch 289045 to 289445. The inter-ranging interruption occurs from the epoch 289045 to 289280, at which all the inter-node ranging measurements are unavailable for MCI and PC aided CP method. From the epoch 289280 to 289445, all the inter-ranging measurements are available for cooperative positioning. It can be seen that both the proposed method and MCI method show satisfied positioning performance during the period when all the inter-node ranges are available. However, the positioning errors of MCI method may increase considerably when the inter-node ranging measurements are missing. Taking the result of the epoch 289265 for an example, the HPE of the MCI method rises to 7.59 m in the case that all the inter-node ranging measurements are removed. The MCI method is equivalent to non-CP method in this case because there is no spatial correlation between $u_1$ and the other three vehicles. Compared to MCI method, the proposed method can still benefit from the positioning information of the other three vehicles by introducing the plane constraint factors, with the figure of HPE declining to 1.23 m at epoch 289265. In short, the proposed CP method can also benefit from cooperative position data even if the inter-node ranging is unavailable.

*5.4 Performance Analysis of Plane Availability Detection and Fault Exclusion*

In the previous analysis, the plane availability detection and fault exclusion are conducted for the proposed method by default. To demonstrate the effectiveness of the fault exclusion (FE) procedures, we make a comparison between the proposed method with FE and the proposed method without FE. It is noted that the inter-node ranging is removed from the proposed CP method so that we can focus on the influence of plane constraints. The control groups are introduced as follows:
- Non-CP: The GNSS standalone positioning method based on RTKLIB.
- PC aided CP without inter-ranging (IR): This is also the CP method proposed in this paper. However, the inter-node ranging measurements are removed from the proposed algorithm. The proposed method becomes a range-free CP method in this case.
- PC aided CP without IR (no FE): This is the proposed method without using IR measurements as well as the fault exclusion procedure.

The CDF of the horizontal and vertical positioning errors for vehicle $u_1$ in three cases are given in Fig. 14 and Fig. 15. The statistic results are given in Table II, including horizontal RMSE, vertical RMSE, CEP95, the maximum HPE and the maximum VPE. It can be seen that there is little performance improvement by introducing plane constraints into CP system without using the plane detection and fault exclusion. Compared to the proposed method with FE, the proposed method without fault exclusion sees a remarkably increase in the positioning errors, with the figure of horizontal RMSE growing from 1.54 m to 1.93 m. Since GNSS multipath effects are inevitable for vehicles traveling in dense urban areas, some positioning results are relatively large, leading to the reliability degradation of the plane constraints. The planes fitted by the positions with large errors are incapable of restraining the positioning solutions, and may even result in the decrease of positioning accuracy.

To show where the fault exclusion occurs during the experiments, the locations where the fault exclusion is conducted are marked in the test route, as shown in Fig. 16. The blue circles represent the positions where the plane detection is passed without fault exclusion. The red circles denote the positions where the plane detection is passed after fault exclusion. The black circles represent the positions where the plane detection cannot be passed even with the help of fault exclusion.

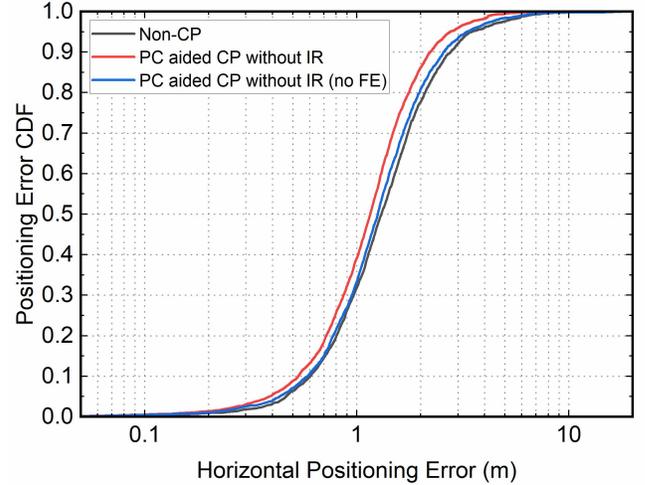

Fig. 14. The CDFs of the horizontal positioning errors for the proposed method with fault exclusion and without fault exclusion.

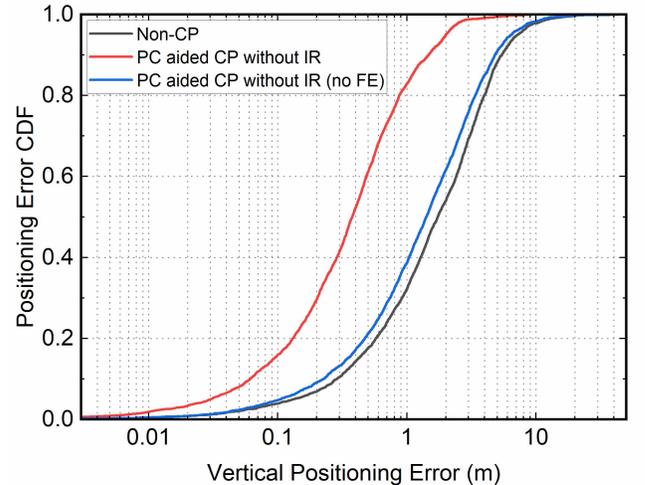

Fig. 15. The CDFs of the vertical positioning errors for the proposed method with fault exclusion and without fault exclusion.

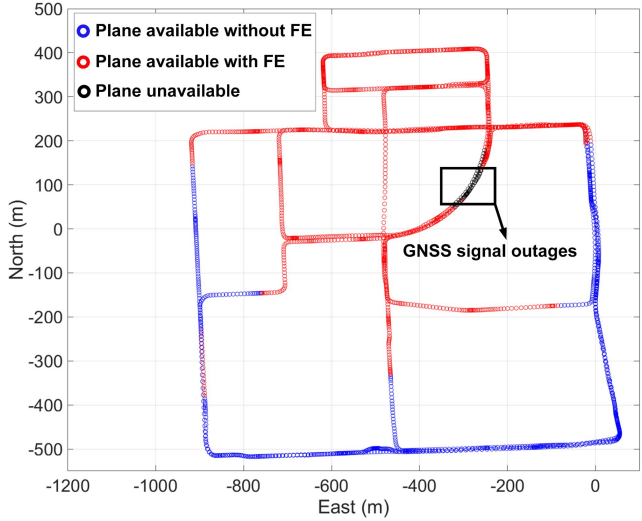

Fig. 16. The locations where the fault exclusion of plane fitting works in the test routes.

The plane constraints are unavailable at the locations of black circles because there are not enough positioning results under the bridge. According to the driving situations depicted in Fig. 5, we find that the fault exclusion is mainly carried out in deep urban canyons where GNSS suffers from severe multipath effects. By using the plane detection and fault exclusion method, the positions with relatively large errors are removed from the plane fitting so as to ensure the reliability of the plane constraints. To sum up, it is necessary to apply the plane availability detection and fault exclusion procedure before using the fitted planes.

TABLE II
STATISTICS FOR POSITIONING PERFORMANCE UNDER DIFFERENT CONFIGURATIONS

| Methods | Horizontal RMSE(m) | Vertical RMSE(m) | CEP95 (m) | Maximum HPE (m) | Maximum VPE (m) |
|---|---|---|---|---|---|
| Non-CP | 1.98 | 3.74 | 3.47 | 15.02 | 36.71 |
| PC aided CP without IR (no FE) | 1.93 | 3.53 | 3.29 | 18.99 | 47.14 |
| PC aided CP without IR | 1.54 | 1.03 | 2.75 | 8.12 | 12.32 |

## VI. CONCLUSION

In this paper, a cooperative positioning method aided by plane constraints is proposed to mitigate the impact of inter-ranging losses. The position-related data from cooperative vehicles are utilized to construct the road plane where the vehicles are traveling. Then the plane parameters can be used to impose the constraints on positioning results. In this way, the position-related data of cooperative vehicles can also be used to conduct cooperative positioning even if the inter-ranging data are missing. A plane availability detection module is designed to determine whether the fitted planes can be used for cooperative positioning. Besides, a fault detection and exclusion method based on RANSAC algorithm is proposed to remove the potential outliers in the plane fitting.

The experimental results show the superiority of the proposed method over the existing CP methods, especially when the inter-vehicle ranging interruptions occur. Compared to other range-based methods, the proposed method can still benefit from the position-related data of the cooperative vehicles by introducing the plane constraint factors. The effectiveness of the plane detection and fault exclusion is also validated by experimental results.

The future work will focus on more complicated road conditions, such as bridges and hilly lands. Theoretically, the plane constraints can also be used on a slope as long as the curvature of the slope is small. We also plan to design a decentralized CP scheme rather than the centralized scheme in this paper.